\documentclass[tablecaption=bottom,wcp]{jmlr} 

\usepackage{booktabs}
\usepackage{bm}
\usepackage[load-configurations=version-1]{siunitx} 

\theorembodyfont{\upshape}
\theoremheaderfont{\scshape}
\theorempostheader{:}
\theoremsep{\newline}

\jmlrproceedings{AABI 2019}{2nd Symposium on Advances in Approximate Bayesian Inference, 2019}

\title[Learning undirected models via query training]{Learning undirected models via query training}

 \author{\Name{Miguel L\'azaro-Gredilla} \Email{miguel@vicarious.com}\\
  \Name{Wolfgang Lehrach} \Email{wolfgang@vicarious.com}\\
  \Name{Dileep George} \Email{dileep@vicarious.com}\\
  \addr Vicarious AI, California}

\begin{document}

\maketitle

\begin{abstract}
Typical amortized inference in variational autoencoders is specialized for a single probabilistic query. Here we propose an inference network architecture that generalizes to unseen probabilistic queries. Instead of an encoder-decoder pair, we can train a single inference network directly from data, using a cost function that is stochastic not only over samples, but also over queries. We can use this network to perform the same inference tasks as we would in an undirected graphical model with hidden variables, without having to deal with the intractable partition function. The results can be mapped to the learning of an actual undirected model, which is a notoriously hard problem. Our network also marginalizes nuisance variables as required.  We show that our approach generalizes to unseen probabilistic queries on also unseen test data, providing fast and flexible inference. Experiments show that this approach outperforms or matches PCD and AdVIL on 9 benchmark datasets.
\end{abstract}


\section{Introduction}
\label{sec:intro}

Learning the parameters of an undirected probabilistic graphical model (PGM) with hidden variables using maximum likelihood (ML) is a notably difficult problem \citep{welling2005learning,kuleshov2017neural,advil}.
When all variables are observed, the range of applicable techniques is broadened \citep{sutton2005piecewise,sutton2006local,sutton2007piecewise,bradley2013learning}, but the problem remains intractable in general. When hidden variables are present, the intractability is twofold: (a) integrating out the hidden variables (also a challenge in directed models) and (b) computing the partition function.
The second problem is generally deemed to be harder \citep{welling2005learning}.

After learning, the probabilistic queries are in most cases not tractable either, so one has to resort to approximations such as belief propagation or variational inference. These approximations operate in the same way regardless of whether the model is directed, and do not need to compute the partition function. In general, ML learning is harder than inference both in directed and undirected models, but even more so in the latter case.

Approximate inference via belief propagation (BP) or variational inference (VI) can be cast as an optimization problem. As such, it rarely has a closed-form solution and is instead solved iteratively, which is computationally intensive. To address this problem, one can use amortized inference. A prime example of this are variational autoencoders \citep{kingma2013auto}: a learned function (typically a neural network) is combined with the reparameterization trick \citep{rezende2014stochastic, titsias2014doubly} to compute the posterior over the hidden variables given the visible ones. Although a variational autoencoder (VAE) performs inference much faster than actual VI optimization, this is not without limitations: \emph{they are specialized to answer a single predefined query}. In contrast, BP and VI answer arbitrary queries, albeit usually need more computation time. 

The end goal of learning the parameters of a PGM is to obtain a model that can answer arbitrary probabilistic queries.
A probabilistic query requests the distribution of a subset of the variables of the model given some (possibly soft) evidence about another subset of variables. This allows, for instance, to train a model on full images and then perform inpainting in a test image in an arbitrary region that was not known at training time.

Since the end goal is to be able to perform arbitrary inference, in this work we suggest to learn a system that is able to answer arbitrary probabilistic queries and avoid ML learning altogether, which completely sidesteps the difficulties associated to the partition function. This puts directed and undirected models on equal footing in terms of usability. To this end, we first unroll inference (we will use BP, but other options are possible) over iterations into a neural network (NN) that outputs the result of an arbitrary query, and then we train said NN to increase its prediction accuracy. At training time we randomize the queries, looking for a consistent parameterization of the NN that \emph{generalizes to new queries}. The hope for existence of such a parameterization comes from BP actually working for arbitrary queries in a graphical model with a single parameterization. We call this approach query training (QT).

\section{Query training (QT)}
The starting point is an unnormalized PGM parameterized by $\theta$. Its probability density can be expressed as  $p({\bm x}; \theta) = p({\bm v}, {\bm h}; \theta) \propto \exp(\phi({\bm v}, {\bm h}; \theta))$ , where ${\bm v}$ are the visible variables available in our data and $\bm{h}$ are the hidden variables. A query is a binary vector $\bm{q}$ of the same dimension as $\bm{v}$ that partitions the visible variables in two subsets: One for which (soft) evidence is available (inputs) and another whose conditional probability we want to estimate (outputs).

\begin{figure}[!h]
  \includegraphics[width=\linewidth]{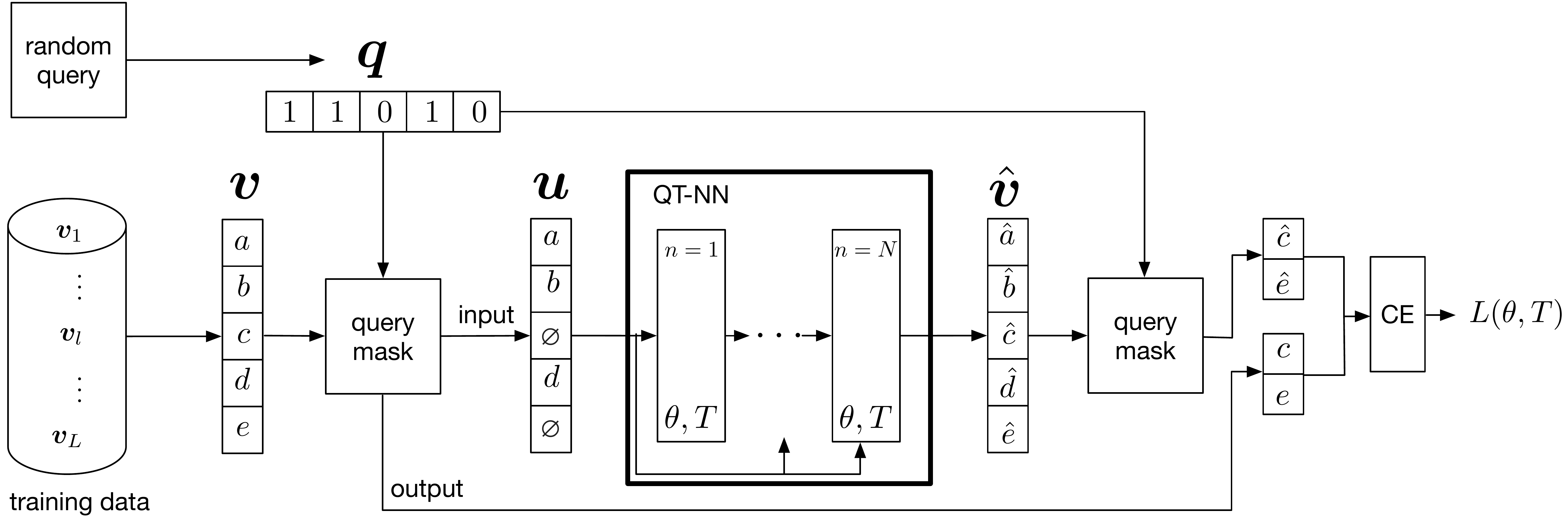}
  \caption{One step of query training. A random sample from the training data is split according to a random query mask in input and output dimensions. The input is processed inside the QT-NN by $N$ identical stages, producing an estimation of the sample. The cross-entropy between the true and estimated outputs is computed.}
  \label{fig:qt}
\end{figure}

\subsection{Training a QT-NN}

The query-trained neural network (QT-NN) follows from specifying a graphical model $\phi({\bm v}, {\bm h}; \theta)$, a temperature $T$ and a number of inference timesteps $N$ over which to run parallel BP. The general equations of the QT-NN are given next in Section \ref{sec:bptonn}, and the equations for the simple case in which the PGM is an RBM is provided in Appendix \ref{sec:rbmcase}.

As depicted in Fig.~\ref{fig:qt}, a QT-NN takes as input a sample $\bm {v}$ from the dataset and a query mask $\bm {q}$. The query $\bm{q}$ blocks the network from accessing the ``output'' variables, and instead only offers access to the ``input'' variables. Which variables are inputs and which ones are outputs is precisely the information that $\bm{q}$ contains. Then the QT-NN produces as output an estimation $\hat{\bm {v}}$ of the whole input sample. Obviously, we only care about how well the network estimates the variables that it did not see at the input. So we measure how well $\hat{\bm {v}}$ matches the correct $\bm {v}$ in terms of cross-entropy (CE), but only for the variables that $\bm{q}$ regards as ``output''.

Taking expectation wrt $\bm {v}$  and $\bm {q}$, we get the loss function that we use to train the QT-NN
$$
\hat{\bm{v}} = \operatorname{QT-NN}(\bm{v}, \bm{q}; \theta, T)
~~~~~~~~
L(\theta, T) = \mathbb{E}_{\bm{v},\bm{q}}[\operatorname{CE}_{\bm{q}}(\bm{v}, \hat{\bm{v}})].
$$

We minimize this loss wrt $\theta, T$ via stochastic gradient descent, sampling from the training data and some query distribution. The number of QT-NN layers $N$ is fixed a priori.

One can think of the QT-NN as a more flexible version of the encoder in a VAE: instead of hardcoding  inference for a single query (normally, hidden variables given visible variables), the QT-NN also takes as input a mask $\bm {q}$ specifying which variables are observed, and provides inference results for unobserved ones. Note that $\bm{h}$ is never observed.

\subsection{Turning BP into a QT-NN}
\label{sec:bptonn}

For a given set of graphical model parameters $\theta$ and temperature $T$ we can write a feed-forward function that approximately resolves arbitrary inference queries by unrolling the parallel BP equations for $N$ iterations.

First, we combine the available evidence $\bm{v}$ and the query $\bm{q}$ into a set of unary factors. Unary factors specify a probability density function over a variable. Therefore, for each dimension inside $\bm{v}$ that $\bm{q}$ labels as ``input'', we provide a (Dirac or Kronecker) delta centered at the value of that dimension. For the ``output'' dimensions and hidden variables $\bm{h}$ we set the unary factor to an uninformative, uniform density. Finally, soft evidence, if present, can be incorporated through the appropriate density function. The result of this process is a unary vector of factors $\bm{u}$ that contains an informative density exclusively about the inputs and whose dimensionality is the sum of the dimensionalities of $\bm{v}$ and $\bm{h}$. Each dimension of $\bm{u}$ will be a real number for binary variables, and a full distribution in the general case.

Once $\bm{v}$ and the query $\bm{q}$ are encoded in $\bm{u}$, we can write down the equations of parallel BP over iterations as an NN with $N$ layers, i.e., the QT-NN. To simplify notation, let us consider a factor graph that contains only pairwise factors. Then the probabilistic predictions of the QT-NN and the messages from each layer to the next can be written as:
\begin{align*}
\hat{\bm{v}}_i = \operatorname{softmax}\Big(\theta_i + \bm{u}_i + \sum_k m_{ki}^{(N)}\Big) &\quad\quad m_{ij}^{(n)} = f_{\theta_{ij}}\Big(\theta_i + \bm{u}_i + \sum_{k\neq j} m_{ki}^{{(n-1)}} ; T\Big) &\quad m_{ij}^{(0)} = 0\\
\text{or equivalently } \quad \hat{\bm{v}} = g_\theta(\bm{m}^{(N)}, \bm{u}) &\quad\quad  \bm{m}^{(n)} = f_\theta(\bm{m}^{(n-1)}, \bm{u}; T) &\quad \bm{m}^{(0)} = 0
\end{align*}

Here $\bm{m}^{(n)}$ collects all the messages\footnote{For a fully connected graph, the number of messages is quadratic in the number of variables, showing the advantage of a sparse connectivity pattern, which can be encoded in the PGM choice.} that exit layer $n-1$ and enter layer $n$. Messages have direction, so $m_{ij}^{(n)}$ is different from $m_{ji}^{(n)}$. Observe how the input term $\bm{u}$ is re-fed at every layer. The output of the network is a belief $\hat{\bm{v}}_i$ for each variable $i$, which is obtained by a softmax in the last layer.  All these equations follow simply from unrolling BP over iterations, with its messages encoded in log-space.

The portion of the parameters $\theta$ relevant to the factor between variables $i$ and $j$ is represented by $\theta_{ij} = \theta_{ji}$, and the portion that only affects variable $i$ is contained in $\theta_i$. Observe that all layers share the same parameters. The functions $f_{\theta_{ij}}(\cdot)$ are directly derived from $\phi(\bm{x};\theta)$ using the BP equations, and therefore inherit its parameters. Finally, parameter $T$ is the ``temperature'' of the message passing, and can be set to $T=1$ to retrieve the standard sum-product belief propagation or to 0 to recover max-product belief revision. Values in-between interpolate between sum-product and max-product and increase the flexibility of the NN. See Appendix \ref{sec:rbmcase} for the precise equations obtained when the PGM is an RBM.

\subsection{Connection with pseudo-likelihood}
If the distribution over queries only contains queries with a single variable assigned as output (and the rest as input), and there are no hidden variables, the above cost function reduces to pseudo-likelihood training \citep{besag1975statistical}. Query training is superior to pseudo-likelihood (PL) in two ways: Firstly, it provides an explicit mechanism for handling hidden variables, and secondly and more importantly, it preserves learning in the face of high correlations in the input data, which results in catastrophic failure when using PL. If two variables $a$ and $b$ are highly correlated, PL will fail to learn the weaker correlation between $a$ and $z$, since $b$ will always be available during training to predict $a$, rendering any correlation with $z$ useless at training time. If at test time we want to predict $a$ from $z$ because $b$ is not available, the prediction will fail. In contrast, query training removes multiple variables from the input, driving the model to better leverage all available sources of information.

\section{Experiments}
\label{sec:experiments}

Early works in learning undirected PGMs relied on contrastive energies \citep{hinton2002training,welling2005learning}. More recent approaches are NVIL \citep{kuleshov2017neural} and AdVIL \citep{advil}, with the latter being regarded as superior. We will use an RBM in our experiments and compare QT with PCD which is very competitive in this setting \citep{tieleman2008training,marlin2010inductive}. We also show results for AdVIL, although it is not necessarily expected to be superior to PCD for this model.

We use exactly the same datasets and preprocessing used in the AdVIL paper, with the same RBM sizes, check \citep{advil} for further details. The random queries are generated by assigning each variable to input or output with 0.5 chance. We report the normalized cross-entropy (NCE), which is the aggregated cross-entropy over the test data, divided by the cross-entropy of a uniform model under the same query (i.e., values below 1.0 mean a better-than-trivial model).

Computing the NCE for QT is as simple as running the trained QT-NN. PCD and AdVIL, however, cannot solve arbitrary inference queries directly and one has to resort to slow Gibbs sampling in the learned model. Alternatively, one can turn the RBM weights learned by this methods into a QT-NN with $T=1$ (essentially, running BP for a fixed number of iterations). We also provide those results as PCD-BP and AdVIL-BP.

For PCD we train for 1000 epochs and cross-validate the learning parameter. For AdVIL we use the code provided by the authors. For QT we unfold BP in $N=10$ layers and use ADAM to learn the weights. The validation set is used to choose the learning rate and for early stopping. We use minibatches of size 500. The $T$ parameter is learned during training. The results are shown in Table \ref{tab:results}.

QT-NN produces significantly better results for most datasets (marked in boldface), showing that it has learned to generalize to new probabilistic queries on unseen data.

\begin{table}[hbtp]
\resizebox{\textwidth}{!}{%
  \begin{tabular}{lcccccccccc}
  \toprule
  Method & \bfseries Adult & \bfseries Conn4 & \bfseries Digits & \bfseries DNA& \bfseries Mushr& \bfseries NIPS& \bfseries OCR& \bfseries RCV1& \bfseries Web \\
  \midrule
AdVIL-BP  & 0.224  & 0.248  & 0.530  & 0.778  & 0.192  & 0.795  & 0.470  & 0.475  & 0.142 \\
AdVIL-Gibbs  & 0.229  & 0.238  & 0.493  & 0.782  & 0.218  & 0.797  & 0.471  & 0.477  & 0.163 \\
PCD-BP  & 0.215  & 0.285  & 0.530  & 0.763  & 0.159  & 0.801  & 0.428  & 0.457  & 0.140 \\
PCD-Gibbs  & 0.218  & 0.288  & 0.516  & 0.765  & 0.159  & 0.804  & 0.427  & 0.458  & 0.144 \\
QT-NN (Ours)  & \textbf{0.167}  & \textbf{0.148}  & \textbf{0.472}  & 0.766  & \textbf{0.124}  & \textbf{0.787}  & \textbf{0.377}  & \textbf{0.452}  & \textbf{0.133} \\
\bottomrule
  \end{tabular}}
  \label{tab:results}
  {\caption{Comparison of QT-NN, PCD and AdVIL. Measurements are NCE, lower is better. See text for details.}}
\end{table}

\section{Discussion and future work}
\label{sec:future}

Query training is a general approach to learn to infer when the inference target is unknown at training time. It offers the following advantages:
1) no need to estimate the partition function or its gradient (the  ``sleep'' phase of other common algorithms);
2) produces an inference network, which can be faster and more accurate than iterative VI or BP because its weights are trained to compensate for the imperfections of approximate inference run for a small number of iterations;
3) arbitrary queries can be solved. In contrast, a VAE is only  trained to infer the posterior over the hidden variables, or some other constant query.

Why would QT-NNs generalize to new queries or scale well? The worry is that only a small fraction of the exponential number of potential queries is seen during training. The \emph{existence} of a single inference network that works reasonably well for many different queries follows from the existence of a single PGM in which BP can approximate inference. The \emph{discoverability} of such a network from limited training data is not guaranteed. However, there is hope for it, since the amount of training data required to adjust the model parameters should scale with the number of these, and not with the number of potential queries. Just like training data should come from the same distribution as test data, the training queries must come from the same distribution the test queries to avoid ``query overfitting''.

In future work we will show how QT can be used in more complex undirected models, such as grid MRFs. Other interesting research avenues are modifications to allow sample generation  and unroll other inference mechanisms, such as VI.


\bibliography{learning_undirected}

\newpage
\appendix

\section{QT-NN equations for the RBM case}
\label{sec:rbmcase}

\begin{figure}[!h]
  \includegraphics[width=\linewidth]{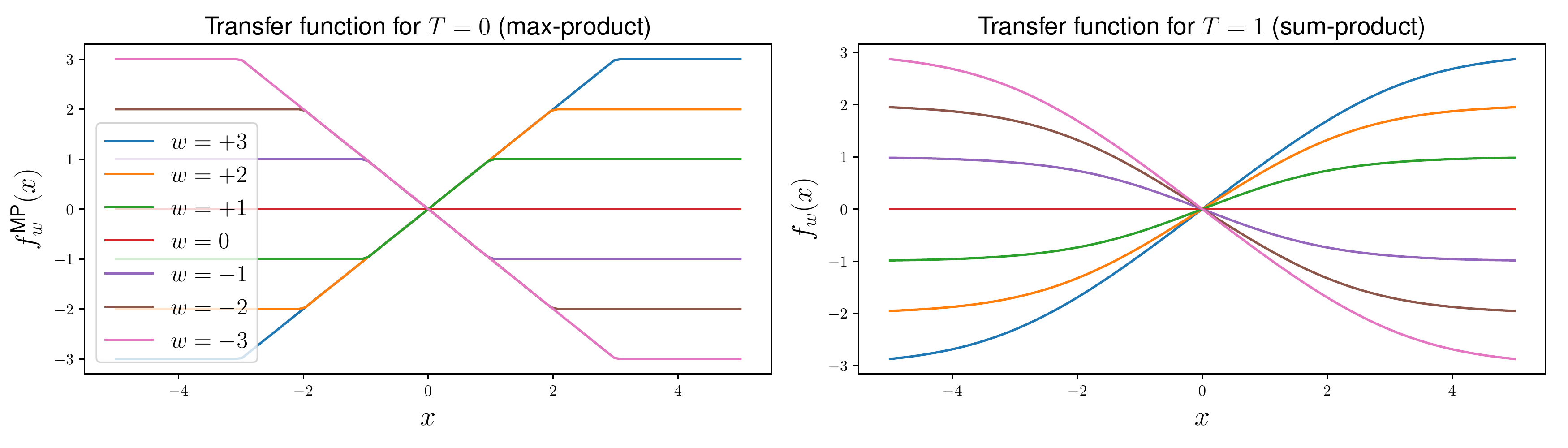}
  \caption{Transfer function for binary pairwise factors (with score 0 for agreement and $-w$ for disagreement) at two temperatures, max-product and sum-product. With this parameterization, setting a weight or input to zero results in zero output.}
  \label{fig:transfer}
\end{figure}

We will consider the simple case in which the underlying PGM is a binary RBM with $H$ hidden units and $V$ visible units. We will use a slightly different parameterization (a linear transformation of the standard one) to simplify the form of the obtained transfer function. Thus, we set $\phi({\bm v}, {\bm h}; \theta) = 2{\bm h}^\top W{\bm v} + {\bm h}^\top (\bm{c}_H-W\bm{1}_V) + {\bm v}^\top (\bm{c}_V-W^\top1_H)$. Then the architecture of the QT-NN is described by the following equations (which simply correspond to unrolling parallel BP over time using messages in logit space):

\begin{align}
\bm{u}_V &= \operatorname{logit}(\bm{v})\circ\bm{q}~~~\text{(unary term for visible units)}\\
\bm{u}_H &= \bm{0}_H~~~\text{(unary term for hidden units)}\\
M_{HV}^{(0)} &= \bm{0}_{HV}~~~\text{(init messages from visible to hidden to 0)}\\
M_{VH}^{(0)} &= \bm{0}_{VH}~~~\text{(init messages from hidden to visible to 0)}\\
M_{HV}^{(n)} &= f_{W^\top}(\bm{u}_V + \bm{c}_V + M_{VH}^{(n-1)} \bm{1}_H - M_{VH})^\top~~~\text{(interlayer connection)}\\
M_{VH}^{(n)} &= f_W(\bm{u}_H + \bm{c}_H + M_{HV}^{(n-1)} \bm{1}_V - M_{HV})^\top~~~\text{(interlayer connection)}\\
\hat{\bm{v}} &= \sigma(\bm{u}_V + \bm{c}_V + M_{VH}^{(N)}\bm{1}_H)~~~\text{(output layer for visible)}\\
\hat{\bm{h}} &= \sigma(\bm{u}_H + \bm{c}_H + M_{HV}^{(N)}\bm{1}_V)~~~\text{(output layer for hidden)}\\,
\end{align}
where
\begin{align}
\sigma(x) &= 1/(1+e^{-x})\\
\operatorname{logit}(x) &= \sigma^{-1}(x) = \log(x) - \log(1-x)\\
f_w^{\text{MP}}(x) &= \operatorname{sign}(w) x|_{-|w|}^{|w|} ~~~\text{($a|_{b}^{c}$ truncates $a$ between $b$ and $c$, Fig.~\ref{fig:transfer} left)}\\
f_w(x) &= f_w^{\text{MP}}(x) + \operatorname{sp}(-|x + w|, T) - \operatorname{sp}(-|x - w|, T) ~~~\text{(Fig.~\ref{fig:transfer} right) }\\
\operatorname{sp}(x, T) &= T \log(1+e^{x/T})~~~\text{(a.k.a. softplus function)}.
\end{align}

Notation clarifications:
\begin{itemize}
  \item We use $\bm{0}_{HV}$ to represent a matrix of zeros of size $H \times V$.
  \item Similarly $\bm{1}_{V}$ represents a matrix of ones of size $V \times 1$.
  \item When any of the above defined scalar functions is used with matrix arguments, the function is applied elementwise.
\end{itemize}

Some observations:
\begin{itemize}
  \item The Hadamard product with $\bm{q}$ effectively removes the information from the elements of $\bm{v}$ not present in the query mask, replacing them with 0, which corresponds to a uniform binary distribution in logit space.
  \item The output of the network is $\hat{\bm{v}}$ and $\hat{\bm{h}}$, the inferred probability of 1 for both the visible and hidden units. The output $\hat{\bm{h}}$ is inferred but actually not used during training.
  \item The computation of $f_w(x)$ as specified above is designed to be numerically robust. It starts by computing $f_w^{\text{MP}}(x)$, which would be the value of $f_w(x)$ for a temperature $T=0$, i.e., max-product message passing, and then performs a correction on top for positive temperatures.
\end{itemize}

\end{document}